\pdfoutput=1

\documentclass[11pt]{article}

\usepackage[preprint]{acl}

\usepackage{times}
\usepackage{latexsym}
\usepackage{booktabs}
\usepackage{multirow}
\usepackage{graphicx}

\usepackage[T1]{fontenc}

\usepackage[utf8]{inputenc}

\usepackage{microtype}

\usepackage{inconsolata}

\usepackage{amsmath}
\usepackage{amssymb}
\usepackage{xspace}
\usepackage{pifont}
\newcommand{\cmark}{\ding{51}}
\newcommand{\xmark}{\ding{55}}

\newcommand{\ensuretext}[1]{#1}

\newif\ifcomments
\ifcomments
\newcommand{\draftcomment}[3]{\ensuretext{\textcolor{#2}{[\ensuretext{\textcolor{#2}{\ensuremath{\textsc{#1}}}} #3]}}}
\else
\newcommand{\draftcomment}[3]{}
\fi

\newcommand{\dataset}{\textsc{SCiFi}\xspace}

\title{Verifiable Generation with Subsentence-Level Fine-Grained Citations}

\author{Shuyang Cao \and Lu Wang \\
  University of Michigan \\
  Ann Arbor, MI \\
  \texttt{\{caoshuy, wangluxy\}@umich.edu}}

\begin{document}
\maketitle

\begin{abstract}

Verifiable generation requires large language models (LLMs) to cite source documents supporting their outputs, thereby improve output transparency and trustworthiness.
Yet, previous work mainly targets the generation of sentence-level citations, lacking specificity about which part of the sentence is backed by which cited source.
This work studies verifiable generation with subsentence-level fine-grained citations to locate the generated content that is supported by the cited sources in a more precise way. 
We first present a dataset, \dataset, comprising 10K Wikipedia paragraphs with subsentence-level citations.\footnote{Our data is available at \url{https://shuyangcao.github.io/projects/subsentence_citation/}.} 
Each paragraph in \dataset is paired with a set of candidate source documents for citation and a query that triggers the generation of the paragraph content.
On \dataset, we then evaluate the performance of state-of-the-a rt LLMs and strategies for processing long documents designed for these models.
Our experiment results reveal key factors that can enhance the quality of citations, including the expansion of the source documents' context to be accessible to the models and the implementation of specialized model tuning.

\end{abstract}
\section{Introduction}

Large language models (LLMs) have demonstrated remarkable capabilities in seeking and synthesizing information from given documents~\cite{touvron2023llama, openai2023gpt4}, and many LLM-powered tools are available to the general public. However, concerns have emerged regarding their outputs' factual accuracy, faithfulness to the source documents, and trustworthiness in general~\cite{zhang2023sirens, peskoff-stewart-2023-credible}.

\begin{figure}[t]
    \centering
    \includegraphics[width=0.48\textwidth]{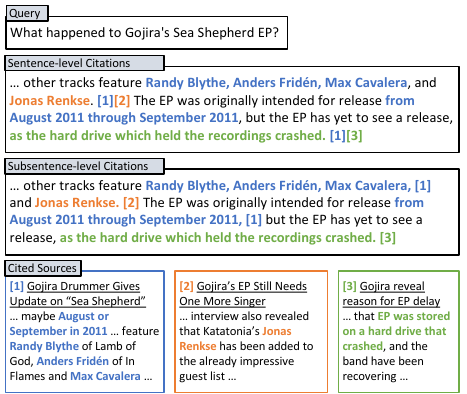}
    \caption{Example of subsentence-level citations. Compared to sentence-level citations, the finer granularity of subsentence-level citations more precisely connect the generated content with the supporting source documents.}
    \label{fig:intro_example}
\end{figure}

To address these concerns, recent research has introduced a new generation paradigm, \textit{verifiable generation}, where LLMs are required to include citations to source documents in the model outputs, to support their statements~\cite{bohnet2023attributed, gao-etal-2023-enabling}.
Verifiable generation enables users to trace the information back to its source and verify its correctness, enhancing the transparency and reliability of models. 
Nonetheless, existing work typically provides sentence-level citations. They are unable to indicate which part of the sentence is supported by each referenced source document, leaving the effort to users to infer the connection between the content and its citation~\cite{liu-etal-2023-evaluating, schuster2023semqa}.
This ambiguity hinders the user's ability to verify the information efficiently and understand the scope of the supporting documents. 

We argue that verifiable generation with a \textit{finer granularity} is critical for further improving the transparency and trustworthiness of LLMs. 
Compared to sentence-level citations, fine-grained citations, such as subsentence-level annotations, allow for more precise localization of the information that is sourced from the referenced document and support easier assessment of its accuracy, as shown in Figure~\ref{fig:intro_example}. 
Furthermore, the importance of fine-grained citations to readers is quantitatively evinced by their prevalence in the popular Wikipedia pages. For instance, pages that are more frequently viewed are more likely to use citations with finer granularity, as demonstrated in Figure~\ref{fig:motivation_quantitative}.

In this work, we aim to investigate LLMs' ability of producing output with fine-grained citations.
Specifically, we focus on subsentence-level citations, which are commonly used in information-rich documents like encyclopedias and research papers.
To facilitate the study, we first collect \textbf{\dataset}, containing \textbf{Sub}sentence-level \textbf{Ci}tation of \textbf{Fi}ne granularity based on 10K paragraphs from Wikipedia, where rich citation information is available. 
For each paragraph in \dataset, documents that are cited in the Wikipedia page where the paragraph belongs to are provided as the citation candidates.
We also create a query based on the content of each paragraph to guide LLMs to generate relevant content.

We benchmark state-of-the-art LLMs on our dataset, including OpenAI GPT ~\cite{openai2023gpt4}, Llama2~\cite{touvron2023llama}, Vicuna~\cite{zheng2023judging}, and Mistral~\cite{jiang2023mistral}. 
To consume the lengthy source documents, we explore three document reading strategies, that respectively target leading context of all source documents, a large portion of the context of all source documents, and full context of selected source documents.
For open-source LLMs, we also examine the effect of fine-tuning with the training samples in our dataset.
For evaluation, we assess the citation behavior, citation quality, and answer quality of model output. We find that: (1) complete source document context improves citation quality in LLMs; (2) larger model sizes increase answer quality but not citation quality; (3) fine-grained citation generation requires supervised fine-tuning.

Our contributions are summarized as follows:
(1) We collect \dataset which consists of 10K queries paired with reference answers that are rich in subsentence-level fine-grained citations to the source documents, enabling training models for verifiable generation with finer granularity.
(2) We analyze performance of state-of-the-art LLMs augmented with various document processing strategies on \dataset, highlighting directions that could advance model development for verifiable generation.

\section{Related Work}

\paragraph{Verifiable Generation.}

Forming in-line citations in verifiable generation challenges LLMs' abilities to ground their generation in source documents and perform accurate attribution.
To teach LLMs to include in-line citations in their outputs, early work fine-tunes LLMs with human written demonstrations~\cite{nakano2022webgpt} or model-generated samples verified by human annotators~\cite{menick2022teaching}, but their privately hosted training data prevents follow-up studies.
The introduction of more capable LLMs~\cite{jiang2023mistral, openai2023gpt4} makes it feasible to prompt LLMs with well-crafted instructions to cite source documents during generation~\cite{gao-etal-2023-enabling}, 
and such behavior has been activated in online systems that are based on LLMs~\cite{liu-etal-2023-evaluating}, though the quality of the generated citations leaves large room for improvement~\cite{malaviya2023expertqa}. 
To enhance the citation quality, recent studies consider fetching source documents that better entail the output content~\cite{li2023llatrieval} or enabling LLMs to refine its outputs~\cite{sun2024verifiable}. 
Nevertheless, existing work largely focuses on verifiable generation with sentence-level citations, without clearly indicating the exact portion in the output that is supported by the source documents. This work, on the other hand, explores verifiable generation with finer granularity.

\paragraph{Fact-based Evaluation.}

Evaluation of outputs produced by LLMs is challenging due to their open-ended nature. Recent work resorts to fact-based evaluation, where an output is first decomposed into independent facts and compared against facts in the reference output~\cite{liu-etal-2023-revisiting}.
To circumvent data collection for fact decomposition model training, LLMs have been instructed to extract facts from the output to be evaluated~\cite{kamoi-etal-2023-wice}.
Furthermore, \citet{min-etal-2023-factscore} leverage LLMs' strong capability of identifying content with similar semantics and design an LLM-based fact comparison module for more accurate evaluation.
While previous work utilizes fact-based framework for evaluation of factual entailment and precision, we assess both answer coverage and citation quality with fact-based evaluation.

\section{\dataset}
\label{sec:dataset}

To benchmark LLMs on verifiable generation with finer granularity, we first collect \dataset, a dataset containing questions that require synthesizing information from multiple source documents to answer and demand more fine-grained citations to precisely locate information supported by different source documents.
\dataset is based on Wikipedia, as Wikipedia articles may densely include citations to support their content.

\begin{figure}
    \centering
    \includegraphics[width=0.45\textwidth]{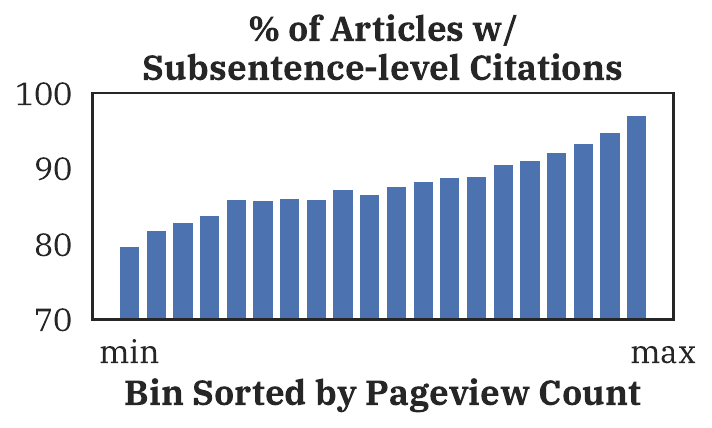}
    \caption{Percentages of Wikipedia articles featuring subsentence-level citations, reported across 20 bins sorted by pageview counts. Popular articles are more likely to include fine-grained citations.}
    \label{fig:motivation_quantitative}
\end{figure}

Before data collection, we examine how prevalent fine-grained citations are used in Wikipedia articles. 
Out of $100$K randomly picked Wikipedia articles, $85\%$ feature subsentence-level citations.
We further sort these articles based on their pageview statistics and divide them into $20$ bins.\footnote{The pageview statistics are obtained via the Wikimedia pageview API: \url{https://wikitech.wikimedia.org/wiki/Analytics/AQS/Pageviews}.}
For each article, we tallied the total number of views over a five-year period to minimize the impact of transient trends.
As can be seen from Figure~\ref{fig:motivation_quantitative}, bins with higher pageview counts show a greater tendency of fine-grained citations, highlighting their importance to readers.

\paragraph{Data Collection.}

For each article, we extract its text content from the Wikipedia dump while keeping track of the positions and referenced sources of citations in the article.
We focus on cited sources that link to downloadable websites and obtain their text content, as it is infeasible to have a uniform process to download and accurately extract content for all types of sources (e.g., images, and PDF files).
The position and meta information of all types of cited sources are still preserved, though, to facilitate other research problems such as identifying citation-worthy content.

We further formulate our dataset into a question-answering dataset by creating queries that serve as constraints of the model generated content.
Queries are crafted for paragraphs with dense citations rather than entire articles, for more appropriate target lengths to support more sophisticated LLM techniques (e.g., in-context learning).
We rank paragraphs with at least $3$ citations based on \textit{citation density}, by dividing the total number of citations by the total number of sentences. From the top $25\%$ of these ranked paragraphs, we randomly sample $10$K paragraphs to promote paragraphs with more fine-grained citations.
For each selected paragraph, we prompt GPT-4 to generate a query asking about the content in the paragraph. 
Having queries to guide and constrain model outputs allows for more robust evaluation of content quality by anchoring to the reference paragraphs.

Eventually, each sample in \dataset contains a reference paragraph, the query generated for the paragraph, and all source documents that are cited by the article where the reference paragraph belongs.
The dataset is split into training and test sets with $9$K and $1$K samples.

\paragraph{Statistics.}

\begin{table}[t]
    \centering
    \small
    \setlength{\tabcolsep}{2pt}
    \begin{tabular}{lrrrc}
    \toprule
        \textbf{Dataset} & \textbf{\# Samples} & \textbf{\# Tokens} & \textbf{Density} & \textbf{Fine-grained?} \\
        \midrule
        \textsc{WiCE} & 1{,}967 & 27.5 & 1.88 &  \xmark \\
        \textsc{HAGRID} & 4{,}532 & 40.5 & 1.23 &  \xmark \\
        \textsc{ExpertQA} & 2{,}177 & 137.2 & 1.27 &  \xmark \\
        \dataset & 10{,}000 & 89.6 & 1.86 & \cmark \\
    \bottomrule
    \end{tabular}
    \caption{Statistics of \dataset and existing datasets. \dataset is larger in size, has a high citation density, and supports fine-grained citations.
    }
    \label{tab:dataset_stat}
\end{table}

We report statistics of \dataset along with recent datasets that involve citations in the outputs, including \textsc{WiCE}~\cite{kamoi-etal-2023-wice}, HAGRID~\cite{kamalloo2023hagrid}, and \textsc{ExpertQA}~\cite{malaviya2023expertqa} in Table~\ref{tab:dataset_stat}.
As each sample in \textsc{WiCE} is a single sentence with citations, for fair comparisons, we compute citation density only for sentences containing at least one citation on all datasets.
Our dataset contains samples with high citation density and moderate target lengths.
More importantly, it includes citations with finer granularity, while others focus on sentence-level citations.
More details of collection and statistics of \dataset are included in Appendix~\ref{appendix:data}.

\section{Experiment Setups}

\paragraph{Task Setup.}

Given a sample in \dataset, the LLM to be tested takes as input the query and candidate source documents to generate a paragraph with fine-grained citations. 
Due to the sheer volume and excessive length of all available source documents, it is impractical to input them into the LLM simultaneously. 
Therefore, we provide the LLM with positive source documents---those cited by the reference paragraph, and $5$ randomly selected negative source documents that are not cited by the reference paragraph.
We shuffle these source documents before feeding them into the LLM.

\paragraph{Models.}

Though we limit the number of source documents input to the model, their concatenation remains lengthy.
To allow the model to process the source documents, we consider
(1) truncating source documents to the same size such that their concatenation can be consumed by the model (\textbf{Truncated});
and (2) providing summaries of source documents to the model (\textbf{Summary}).
Additionally, we design a \textbf{two-stage} framework that iteratively selects the documents to be covered in next sentence by reading their summaries and writes the next sentence by consuming the original text of the selected documents.
This selection process reduces the size of the document pool for the generation phase, thereby allowing the inclusion of more complete document context.

We examine the efficacy of both proprietary and open-source backbone models.
As for proprietary models, we test GPT-3.5 with 16K context length and GPT-4 with 8K context length.
The 7B and 13B variants of Llama2, 7B variant of Vicuna, and 7B variant of Mistral are chosen for the open-source backbones. We use their RLHF-tuned version for all open-source LLMs.
Besides \textbf{in-context learning} that is used for all experimented models, we additionally perform \textbf{supervised fine-tuning} for open-source models with $4{,}000$ samples in the training set of \dataset to investigate the effect of fine-tuning.

\paragraph{Evaluation Metrics.}

We target the assessment of LLMs' ability to produce subsentence-level fine-grained citations, precisely cite the supporting documents, and cover sufficient information for answering the question.
We first calculate the \textbf{citation density} of model outputs, which is the average number of citations in each output sentence.

For citation quality, we follow \citet{ais} and measure how well the output statements entail the cited sources.
Unlike sentence-level citations which can be paired with the entire sentence for entailment assessment, evaluation of fine-grained citations requires segmenting the sentence and mapping citations to the sentence portions they support.
Inspired by \textbf{fact-level entailment}~\cite{min-etal-2023-factscore}, we decompose model outputs into individual facts and pair citations with facts using heuristics (Appendix~\ref{appendix:experiment}).
Following \citet{gao-etal-2023-enabling}, we run an off-the-shelf entailment model~\cite{honovich-etal-2022-true-evaluating} to obtain the entailment levels between generated citations and facts.

Fact-level evaluation is also applied to assess answer quality. For each fact in the reference paragraph, we check if it entails the model output. The aggregation of the fact-level entailment scores reflects the \textbf{coverage} of reference facts (i.e., recall).

\section{Results}

\begin{table}[t]
    \centering
    \small
    \setlength{\tabcolsep}{3pt}
    \begin{tabular}{lcccc}
    \toprule
        \textbf{Strategy} & \textbf{Density} & \textbf{Density (sub)} & \textbf{Citation Ent.}  & \textbf{Cover.} \\
        \midrule
        \multicolumn{5}{l}{\textit{GPT-3.5}} \\
        Truncated & 0.57 & 0.07 & 19.57 & \underline{22.53} \\
        Summary & \underline{0.59} & \underline{0.09} & 29.75 & 22.08 \\
        Two-stage & 0.53 & 0.03 & \underline{35.30} & 19.97 \\
        \midrule
        \multicolumn{5}{l}{\textit{GPT-4}} \\
        Truncated & 0.68 & 0.20 & 39.87 & \underline{\textbf{25.86}} \\
        Summary & 0.84 & \underline{0.26} & 47.00 & 24.53 \\
        Two-stage & \underline{\textbf{1.15}} & 0.25 & \underline{\textbf{58.56}} & 21.60 \\
        \midrule
        \multicolumn{5}{l}{\textit{Llama2-7B}}  \\
        Truncated & \underline{0.57} & \underline{0.17} & 24.03 & 18.59 \\
        Summary & 0.53 & 0.13 & \underline{30.06} & \underline{19.82} \\
        \midrule
        \multicolumn{5}{l}{\textit{Llama2-13B}} \\
        Truncated & \underline{0.51} & \underline{0.17} & 17.87 & 20.20 \\
        Summary & 0.49 & \underline{0.17} & \underline{21.23} & \underline{21.91} \\
        \midrule
        \multicolumn{5}{l}{\textit{Vicuna-7B}} \\
        Truncated & \underline{0.80} & \underline{0.12} & 27.48 & 16.76 \\
        Summary & 0.64 & 0.09 & \underline{30.10} & \underline{18.69} \\
        \midrule
        \multicolumn{5}{l}{\textit{Mistral-7B}} \\
        Truncated & 1.06 & \underline{\textbf{0.31}} & 48.99 & \underline{20.23} \\
        Summary & \underline{1.07} & \underline{\textbf{0.31}} & \underline{49.42} & 20.20 \\
    \bottomrule
    \end{tabular}
    \caption{Evaluation results of outputs produced by different strategies and backbone models using in-context learning. (sub): subsentence-level; Ent.: entailment; Cover.: coverage. The best score per metric is in \textbf{bold}, while the best strategy per backbone model is \underline{underlined}. Overall, document reading strategies that provide more complete context yield better citation quality.}
    \label{tab:result_strategy_icl}
\end{table}

\textbf{Source documents with more complete context benefit citation quality}, as indicated by the fact-level entailment scores of citations produced by different document reading strategies with in-context learning (Table~\ref{tab:result_strategy_icl}). 
Summaries can inform models of the major content in the source documents, while truncation only exposes leading content and prevents accurate connection between generated content and supporting documents, thus consistently yielding worse citation quality.
Two-stage generation allows for the most complete document context, boosting the citation accuracy, yet its effectiveness relies on strong instruction-following capabilities of the LLMs.\footnote{The two-stage strategy is only adopted by the OpenAI GPT families, as other models could not consistently follow the output format designed for the strategy, resulting in invalid results.}

\textbf{Increasing model sizes promotes answer quality.}
Across different backbone models and strategies, the coverage of reference facts increases after switching to a larger model within the same family, though larger model sizes do not guarantee an enhancement in citation quality.
This reveals that the pre-training designs of different backbone LLMs might all aim for stronger question-answering capabilities, but assign varying significance to their citation and attribution capabilities.

\begin{table}[t]
    \centering
    \small
    \setlength{\tabcolsep}{3pt}
    \begin{tabular}{lcccc}
    \toprule
        \textbf{Strategy} & \textbf{Density} & \textbf{Density (sub)} & \textbf{Citation Ent.}  & \textbf{Cover.} \\
        \midrule
        \multicolumn{5}{l}{\textit{Llama-13B}} \\
        Truncated & +0.78 & +0.27 & +14.22 & -2.86 \\
        Summary & +0.81 & +0.29 & +7.94 & -5.06 \\
        \midrule
        \multicolumn{5}{l}{\textit{Vicuna-7B}} \\
        Truncated & +0.81 & +0.60 & +1.50 & -1.42 \\
        Summary & +0.89 & +0.54 & -0.87 & -2.31 \\
    \bottomrule
    \end{tabular}
    \caption{Improvement of performance after supervised fine-tuning. Negative numbers indicate drops in performance. Supervised fine-tuning encourages model to produce more subsentence-level citations, though not always for citation quality.}
    \label{tab:result_icl_sft}
\end{table}

\textbf{Generation of fine-grained citations requires additional training.}
We observe that the density of fine-grained citations generated by the same backbone LLMs remains stable across different document reading strategies.
By contrast, models generate substantially more fine-grained citations after supervised fine-tuning, as shown in Table~\ref{tab:result_icl_sft}.
However, the effectiveness of supervised fine-tuning in enhancing citation quality varies across models.
We think that supervised fine-tuning encourages LLMs' behaviors of generating fine-grained citations. Yet, the development of LLMs' abilities to correctly link sentence parts with supporting documents requires more specialized and sophisticated training procedure, which highlights the challenge of this task. Future directions may include design builtin citation or attribution mechanisms during LLM pretraining~\cite{khalifa2024source}.

\section{Conclusions}

We study verifiable generation with subsentence-level fine-grained citations.
\dataset, a benchmark containing $10$K subsentence-level citation-rich paragraphs together with candidate cited sources and queries, is collected to support the training and evaluation of models on this task.
On \dataset, experiments with state-of-the-art LLMs and various processing strategies demonstrate the importance of source document context and training with citation-rich data.

\section*{Acknowledgments}
This work is supported in part by the National Science Foundation through grant IIS-2046016. Shuyang Cao is supported by a Bloomberg Data Science Ph.D. Fellowship. We thank ARR reviewers for their feedback.
\section*{Limitations}

In our experiments, we employ three strategies for LLMs to handle lengthy source documents and observe improved performance when the strategy enables a more comprehensive context.
Yet, more sophisticated strategies can be designed, and techniques that expand input limits of LLMs can be explored, which could potentially lead to higher performance in our benchmark study.

When evaluating the attribution quality, we leverage an existing entailment model tailored for assessing the entailment relation between short passages. However, we use it to measure the entailment relation between an extracted fact and a long source document.
Although we follow the technique in previous work (see Appendix~\ref{appendix:evaluation_metrics}) to extend the application of the off-the-shelf entailment models to long documents, more accurate evaluation can be achieved by developing entailment models specialized for long documents.

\section*{Ethical Considerations}

Our benchmark enables the evaluation of LLMs’ ability to generate subsentence-level citations. 
With subsentence-level citations, LLM developers are able to present LLM outputs that contain precise pointers directing users to supporting sources of different parts of output sentences.
While this would enhance user trust in LLMs, it is worth noting that our dataset comprises texts that are formally written in Wikipedia and the candidate supporting documents are from reliable online sources. 
An LLM with outstanding performance on our dataset might cite documents with fake facts if the candidate documents are from unreliable sources, which might further propagate incorrect information. 
Users of our benchmark should also consider the reliability of their candidate supporting documents when examining the reliability of LLM-based applications.

\bibliography{custom}

\appendix

\section{Details of Data Collection}
\label{appendix:data}

\subsection{Wikipedia Article and Cited Source Collection}

Wikipedia paragraphs in \dataset are extracted from the Aug 20, 2021 Wikipedia dump.
For each source that links to a downloadable website, we retrieve its HTML file from Internet Archive\footnote{\url{http://web.archive.org/}} and extract its metadata and text content using Trafilatura.\footnote{\url{https://github.com/adbar/trafilatura}}

A sample example is shown in Table~\ref{tab:example_sample}.

Our dataset will be made publicly available under the CC BY 4.0 license.\footnote{\url{https://creativecommons.org/licenses/by/4.0/}}

\subsection{Topic Distribution}

We use the topic model provided by Wikimedia\footnote{\url{https://meta.wikimedia.org/wiki/Machine_learning_models/Production/English_Wikipedia_article_topic}} to determine the topic of the page from which each paragraph in \dataset is extracted. 
For each page, we select its top 3 predicted topics, considering that a single page could cover multiple topics. 
Top 10 topics are presented in the Table~\ref{tab:topic_distribution}.
Paragraphs in our dataset come from pages of diverse topics.

\begin{table}[h]
    \centering
    \small
    \begin{tabular}{lc}
    \toprule
        \textbf{Topic} & \textbf{Percentage} \\
        \midrule
        STEM.STEM & 39.45 \\
        Culture.Media.Media & 23.42 \\
        Geography.Regions.Europe & 17.77 \\
        Culture.Biography & 16.97 \\
        STEM.Technology & 12.95 \\
        Geography.North America & 12.54 \\
        Geography.Asia & 9.13 \\
        History and Society.Politics and Government & 7.78 \\
        Culture.Literature & 6.01 \\
        Culture.Philosophy and Religion & 5.67 \\
         \bottomrule
    \end{tabular}
    \caption{Top 10 topics covered by the pages where samples in \dataset are extracted.}
    \label{tab:topic_distribution}
\end{table}

\subsection{Additional Statistics}

\dataset has 39292 sentences in total, $30.1\%$ of which have more than one citation.
As subsentence-level citations frequently occur at the end of clauses marked with punctuation, we also check the percentage of subsentence-level citations in \dataset that are not attached to punctuation. 
We find that $30.2\%$ of subsentence-level citations are not located around punctuation, indicating a decent level of diversity in subsentence-level citations.

\subsection{Prompt for Query Generation}

We use GPT-4 to create query for each paragraph in \dataset. The prompt we use is shown in Table~\ref{tab:prompt_query_generatioon}.

\begin{table*}[t]
    \centering
    \begin{tabular}{p{0.97\textwidth}}
    \toprule
    \texttt{Read the following paragraph from a Wikipedia page and create a question whose answer covers most of the information in the paragraph. The section title and page title where the paragraph is found are also provided (if any). Try not to create a compound question.} \\
    \texttt{Page Title: \{page\}} \\
    \texttt{Section Title: \{section\}} \\
    \texttt{Paragraph: \{paragraph\}} \\
    \texttt{Question:} \\
    \bottomrule
    \end{tabular}
    \caption{Prompt for query generation.}
    \label{tab:prompt_query_generatioon}
\end{table*}

\section{Additional Results}

\begin{table}[t]
    \centering
    \small
    \setlength{\tabcolsep}{3pt}
    \begin{tabular}{lcccc}
    \toprule
        \textbf{Strategy} & \textbf{Density} & \textbf{Density (sub)} & \textbf{Citation Ent.}  & \textbf{Cover.} \\
        \midrule
        \multicolumn{5}{l}{\textit{GPT-3.5}} \\
        Truncated & 0.49 & \underline{0.06} & 22.68 & 22.53 \\
        Summary & 0.52 & 0.05 & 25.60 & \underline{24.25} \\
        Two-stage & \underline{0.93} & 0.05 & \underline{55.14} & 19.43 \\
        \midrule
        \multicolumn{5}{l}{\textit{GPT-4}} \\
        Truncated & 0.66 & 0.19 & 38.89 & \underline{\textbf{29.85}} \\
        Summary & 0.86 & 0.25 & 47.25 & 24.44 \\
        Two-stage & \underline{\textbf{1.37}} & \underline{\textbf{0.28}} & \underline{\textbf{64.11}} & 20.91 \\
        \midrule
        \multicolumn{5}{l}{\textit{Llama2-7B}}  \\
        Truncated & 0.49 & \underline{0.10} & 26.71 & \underline{23.76} \\
        Summary & \underline{0.50} & \underline{0.10} & \underline{31.16} & 23.42 \\
        \midrule
        \multicolumn{5}{l}{\textit{Llama2-13B}} \\
        Truncated & 0.46 & 0.13 & 21.48 & \underline{25.96} \\
        Summary & \underline{0.48} & \underline{0.15} & \underline{24.01} & 25.24 \\
    \bottomrule
    \end{tabular}
    \caption{Evaluation results of outputs produced by different strategies and backbone models using in-context learning in the oracle setup. (sub): subsentence-level; Ent.: entailment; Cover.: coverage. The best score of each metric among \textbf{bolded.}, while the best strategy for each backbone model is \underline{underlined}.}
    \label{tab:result_oracle}
\end{table}

We additionally test the model performance in the oracle setup, where only positive source documents (i.e., those cited by the reference paragraph) are fed into the LLM (Table~\ref{tab:result_oracle}).
In the oracle setup, trends of the results are similar to those in the regular setup, with all models tending to produce answers of high quality while maintaining the citation quality.
This indicates that removing source documents irrelevant to the query offers minimal help to LLMs for verifiable generation.
\section{Details of Experiment Setups}
\label{appendix:experiment}

\subsection{Model Prompting}

We use 2-shot examples for all experiments with in-context learning.
Prompts for the \textbf{Truncated} and \textbf{Summary} strategy are shown in Table~\ref{tab:prompt_truncated} and~\ref{tab:prompt_summary}.
Summaries of the source documents are generated by GPT-3.5 with 16K context length using the prompt in Table~\ref{tab:prompt_summary_generation}.

In the \textbf{Two-stage} strategy, the LLM is given two different prompts to perform document selection and answer sentence generation.
When selecting source documents, the LLM is informed of the current answer and all its previous selections, as shown in Table~\ref{tab:prompt_two_stage_selection}.
When generating the next answer sentence, the LLM is provided the current answer and the original text of the selected source documents (Table~\ref{tab:prompt_two_stage_generation}).

All models are only prompted once due to the API cost.

\subsection{Supervised Fine-tuning}

We use \texttt{LLaMA-Factory}\footnote{\url{https://github.com/hiyouga/LLaMA-Factory/}} for fine-tuning Llama2 models in our experiments.
We adopt LoRA~\cite{hu2021lora} and fine-tune the model for 3 epochs with a learning rate of 5e-5 and a batch size of 16.
All LoRA-compatible projection layers are tuned, with a rank of 32 and a $\alpha$ of 64.
Training of each model is conducted using 2 Nvidia A40 GPUs and takes 4 hours to complete.

\subsection{Evaluation Metrics}
\label{appendix:evaluation_metrics}

We leverage fact-level entailment to evaluate the citation quality and answer quality.
We follow the prompts in previous work~\cite{min-etal-2023-factscore, kamoi-etal-2023-wice} and use GPT-3.5 to conduct fact decomposition for each output sentence separately.

To map citations to an extracted fact, we first map the extracted fact back to a segment in the original sentence.
For accurate mapping, we again use GPT-3.5 to identify segments in the original sentence that best represents the extracted fact, with the prompt in Table~\ref{tab:prompt_fact_mapping}.
We then rank generated citations based on their distances to the sentence segment associated with the extracted fact. 
If two citations have the same distance to the end of the sentence segment, the one after the sentence segment is ranked higher, as we hypothesize that the citation supporting a fact is likely to occur after its completion in the sentence.
The top-ranking citation is mapped to the extracted fact.

When evaluating the entailment level, directly pairing the cited source document with the extracted fact is infeasible due to the length of the cited document.
Following \citet{kamoi-etal-2023-wice}, we divide the cited document into chunks of 256 tokens, calculate the extracted fact's entailment level against each chunk, and take highest entailment level as the final entailment score.

\begin{table*}[t]
    \centering
    \begin{tabular}{p{0.97\textwidth}}
    \toprule
    \texttt{Instruction: Write an accurate, engaging, and concise answer for the given question using only the provided search results (some of which might be irrelevant) and cite them properly. Use an unbiased and journalistic tone. Always cite after the completion of each individual fact in the answer. Facts might be completed in the middle of a sentence.} \\
    \\
    \texttt{Question: \{query\}} \\
    \\
    \texttt{Document [1] (Title: \{document1\_title\})} \\
    \texttt{\{truncated\_document1\_text\}} \\
    \\
    ... \\
    \\
    \texttt{Document [N] (Title: \{documentN\_title\})} \\
    \texttt{\{truncated\_documentN\_text\}} \\
    \\
    \texttt{Answer:} \\
    \bottomrule
    \end{tabular}
    \caption{Prompt for generation with the \textbf{Truncated} strategy.}
    \label{tab:prompt_truncated}
\end{table*}

\begin{table*}[t]
    \centering
    \begin{tabular}{p{0.97\textwidth}}
    \toprule
    \texttt{Summarize the following document within 100 words. Try to keep all the important dates, numbers, and names.} \\
    \texttt{Title: \{title\}} \\
    \texttt{Text: \{text\}} \\
    \texttt{Summary:} \\
    \bottomrule
    \end{tabular}
    \caption{Prompt for article summary generation.}
    \label{tab:prompt_summary_generation}
\end{table*}

\begin{table*}[t]
    \centering
    \begin{tabular}{p{0.97\textwidth}}
    \toprule
    \texttt{Instruction: Write an accurate, engaging, and concise answer for the given question using only the provided search results (some of which might be irrelevant) and cite them properly. You are provided summaries of the search results, rather than the original search results. Use an unbiased and journalistic tone. Always cite after the completion of each individual fact in the answer. Facts might be completed in the middle of a sentence.} \\
    \\
    \texttt{Question: \{query\}} \\
    \\
    \texttt{Document [1] (Title: \{document1\_title\})} \\
    \texttt{\{summary\_document1\_text\}} \\
    \\
    ... \\
    \\
    \texttt{Document [N] (Title: \{documentN\_title\})} \\
    \texttt{\{summary\_documentN\_text\}} \\
    \\
    \texttt{Answer:} \\
    \bottomrule
    \end{tabular}
    \caption{Prompt for generation with the \textbf{Summary} strategy.}
    \label{tab:prompt_summary}
\end{table*}

\begin{table*}[t]
    \centering
    \begin{tabular}{p{0.97\textwidth}}
    \toprule
    \texttt{Instruction: Write an accurate, engaging, and concise answer for the given question using only the provided search results (some of which might be irrelevant) and cite them properly. You are provided summaries of the search results, rather than the original search results. Answer the question sentence by sentence. Now, given the empty or already written answer, choose which document(s) in the search results to use for the next sentence of the answer. You can also decide to stop the answer with [STOP] if you think the answer is complete.} \\
    \\
    \texttt{Question: \{query\}} \\
    \\
    \texttt{Document [1] (Title: \{document1\_title\})} \\
    \texttt{\{summary\_document1\_text\}} \\
    \\
    ... \\
    \\
    \texttt{Document [N] (Title: \{documentN\_title\})} \\
    \texttt{\{summary\_documentN\_text\}} \\
    \\
    \texttt{Written Answer Sentences: \{current\_answer\_iter1\}} \\
    \texttt{Chosen Documents: \{chosen\_document\_iter1\}} \\
    \\
    ... \\
    \\
    \texttt{Written Answer Sentences: \{current\_answer\_iterN\}} \\
    \texttt{Chosen Documents:} \\
    \bottomrule
    \end{tabular}
    \caption{Prompt for the selection phase of the \textbf{Two-stage} strategy.}
    \label{tab:prompt_two_stage_selection}
\end{table*}

\begin{table*}[t]
    \centering
    \begin{tabular}{p{0.97\textwidth}}
    \toprule
    \texttt{Instruction: Write an accurate, engaging, and concise answer for the given question using only the provided search results (some of which might be irrelevant) and cite them properly. Answer the question sentence by sentence, and the already written answer sentences are given. Now, write the next sentence of the answer. Use an unbiased and journalistic tone. Always cite after the completion of each individual fact in the answer. Facts might be completed in the middle of a sentence.} \\
    \\
    \texttt{Question: \{query\}} \\
    \\
    \texttt{Document [\{selected\_document1\_index\}] (Title: \{selected\_document1\_title\})} \\
    \texttt{\{selected\_document1\_text\}} \\
    \\
    ... \\
    \\
    \texttt{Document [\{selected\_documentN\_index\}] (Title: \{selected\_documentN\_title\})} \\
    \texttt{\{selected\_documentN\_text\}} \\
    \\
    \texttt{Written Answer Sentences: \{current\_answer\}} \\
    \texttt{Next Sentence:} \\
    \bottomrule
    \end{tabular}
    \caption{Prompt for the generation phase of the \textbf{Two-stage} strategy.}
    \label{tab:prompt_two_stage_generation}
\end{table*}

\begin{table*}[t]
    \centering
    \begin{tabular}{p{0.97\textwidth}}
    \toprule
    \texttt{Find the shortest possible segment in the sentence that reflects the claim. The segment must be a contiguous substring of the sentence.} \\
    \texttt{Sentence: He made his acting debut in the film The Moon is the Sun’s Dream (1992), and continued to appear in small and supporting roles throughout the 1990s.} \\
    \texttt{Claim: He made his acting debut in 1992.} \\
    \texttt{Segment: 1992} \\
    \\
    \texttt{Find the shortest possible segment in the sentence that reflects the claim. The segment must be a contiguous substring of the sentence.} \\
    \texttt{Sentence: In 1963, Collins became one of the third group of astronauts selected by NASA and he served as the back-up Command Module Pilot for the Gemini 7 mission.} \\
    \texttt{Claim: Collins became one of the third group of astronauts selected by NASA in 1963.} \\
    \texttt{Segment: In 1963, Collins became one of the third group of astronauts selected by NASA} \\
    \\
    \texttt{Find the shortest possible segment in the sentence that reflects the claim. The segment must be a contiguous substring of the sentence.} \\
    \texttt{Sentence: A previous six time winner of the Nations' Cup, Sebastian Vettel became Champion of Champions for the first time, defeating Tom Kristensen, who made the final for the fourth time, 2–0.} \\
    \texttt{Claim: Sebastian Vettel is a previous six-time winner of the Nations' Cup.} \\
    \texttt{Segment: A previous six time winner of the Nations' Cup} \\
    \\
    \texttt{Find the shortest possible segment in the sentence that reflects the claim. The segment must be a contiguous substring of the sentence.} \\
    \texttt{Sentence: A previous six time winner of the Nations' Cup, Sebastian Vettel became Champion of Champions for the first time, defeating Tom Kristensen, who made the final for the fourth time, 2–0.} \\
    \texttt{Claim: Tom Kristensen made the final for the fourth time.} \\
    \texttt{Segment: Tom Kristensen, who made the final for the fourth time} \\
    \\
    \texttt{Find the shortest possible segment in the sentence that reflects the claim. The segment must be a contiguous substring of the sentence.} \\
    \texttt{Sentence: \{sentence\}} \\
    \texttt{Claim: \{fact\}} \\
    \texttt{Segment:} \\
    \bottomrule
    \end{tabular}
    \caption{Prompt for fact mapping.}
    \label{tab:prompt_fact_mapping}
\end{table*}

\begin{table*}[t]
    \centering
    \begin{tabular}{p{0.97\textwidth}}
    \toprule
    \texttt{Question: What were the circumstances and details of Richard Blumenthal's military service during the Vietnam War?} \\
    \\
    \texttt{Reference: Blumenthal received five draft deferments during the Vietnam War, [5] first educational deferments, then deferments based on his occupation. [1] With part-time service in the reserves or National Guard generally regarded as an alternative for those wishing to avoid service in Vietnam, in April 1970 Blumenthal enlisted in the United States Marine Corps Reserve. He served in units in Washington, D.C., and Connecticut from 1970 to 1976, [2] attaining the rank of sergeant. [4]} \\
    \\
    \texttt{Document [1] (Title: Dick Blumenthal, Reporting for Duty)} \\
    \texttt{Perhaps John Kerry, the former junior senator from Massachusetts, did not serve as heroically in Vietnam as he would like us to think. Certainly he wasn't there for very long. But at least he put in an appearance. The same can't be said of Sgt. Richard Blumenthal, a fellow Democrat and the attorney general of Connecticut, who is seeking to become that state's junior senator ...} \\
    \\
    \texttt{Document [2] (Title: Blumenthal an easy victor)} \\
    \texttt{HARTFORD -- Democratic delegates overwhelmingly and unsurprisingly nominated Attorney General Richard Blumenthal on Friday night to run for the U.S. Senate seat that will be vacated by U.S. Sen. Chris Dodd ...} \\
    \\
    \texttt{Document [3] (Title: David Blumenthal, M.D., M.P.P.)} \\
    \texttt{David Blumenthal, M.D., M.P.P., is president of The Commonwealth Fund, a national philanthropy engaged in independent research on health and social policy issues ...} \\
    \\
    \texttt{Document [4] (Title: Senator Blumenthal honored at Yale Graduate School diversity conference)} \\
    \texttt{The ninth annual Bouchet Leadership Conference on Diversity in Graduate Education took place at Yale March 30-31. The focus of this year’s conference was “Determining the Future of Diversity Discussions.” U.S. Senator Richard Blumenthal ’73 J.D. (D-CT), who received this year’s Bouchet Leadership Award, at the conference, delivered the keynote address ...} \\
    \\
    \texttt{Document [5] (Title: Senate hopeful Richard Blumenthal addresses report he lied about Vietnam record)} \\
    \texttt{Connecticut Attorney General Richard Blumenthal (D) was alternately apologetic and defiant Tuesday as he battled to deflect a potentially devastating blow to his Senate campaign: an accusation that he had exaggerated his military service record ...} \\
    \\
    \texttt{Document [...] (Title: ...)} \\
    \texttt{...} \\
    \bottomrule
    \end{tabular}
    \caption{Sample of \dataset. First paragraphs of the first 5 documents in the candidate document pool are shown.}
    \label{tab:example_sample}
\end{table*}

\end{document}